\documentclass{article}
\usepackage{ijcai22}
\usepackage{times}
\usepackage{soul}
\usepackage{url}
\usepackage{multirow}
\usepackage[hidelinks]{hyperref}
\usepackage[utf8]{inputenc}
\usepackage[small]{caption}
\usepackage{graphicx}
\usepackage{amsmath}
\usepackage{amsthm}
\usepackage{booktabs}
\usepackage{algorithm}
\usepackage{algorithmic}
\title{MixKG: Mixing for harder negative samples in knowledge graph}
\author{
    Feihu Che, Guohua Yang, Pengpeng Shao, Dawei Zhang, Jianhua Tao
}

\begin{document}

\maketitle

\begin{abstract}
Knowledge graph embedding~(KGE) aims to represent entities and relations into low-dimensional vectors for many real-world applications. The representations of entities and relations are learned via contrasting the positive and negative triplets. Thus, high-quality negative samples are extremely important in KGE. However, the present KGE models either rely on simple negative sampling methods, which makes it difficult to obtain informative negative triplets; or employ complex adversarial methods, which requires more training data and strategies. In addition, these methods can only construct negative triplets using the existing entities, which limits the potential to explore harder negative triplets. To address these issues, we adopt mixing operation in generating harder negative samples for knowledge graphs and introduce an inexpensive but effective method called MixKG. 
Technically, MixKG first proposes two kinds of criteria to filter hard negative triplets among the sampled negatives:  based on scoring function and based on correct entity similarity. Then, MixKG synthesizes harder negative samples via the convex combinations of the paired selected hard negatives. 
Experiments on two public datasets and four classical KGE methods show MixKG is superior to previous negative sampling algorithms.
\end{abstract}
\section{Introduction}

Knowledge graphs~(KGs)  store information in the form of factual triplets~(\emph{h}, \emph{r}, \emph{t}), where each triplet represents a particular relation~(\emph{r})  between head entity~(\emph{h}) and tail entity~(\emph{t}). Recently, with the help of KGs, many related applications have gained performance improvements, such as question answering~\cite{yao2014information}, information retrieval~\cite{xiong2017explicit} and machine reading~\cite{hao2017end}. To benefit more applications, there exist many famous KGs, such as Freebase~\cite{bollacker2008freebase}, Yago3~\cite{mahdisoltani2014yago3} and WordNet~\cite{miller1995wordnet}. 

However, directly using factual triplets in KGs is known to be difficult~\cite{zhang2019nscaching}. The current popular methods are to embed entities and relations in the KGs into a vector space~\cite{bordes2013translating,nguyen2017overview,sun2019rotate}. Recently, some researchers try to employ graph embedding methods for KG embeddings~\cite{shang2019end,schlichtkrull2018modeling}. The great advantage of these models is that they map entities and relations in KGs into a low-dimensional vector space while preserving the information about graph structure. Therefore, they have shown promising performance in KG related tasks, such as link prediction~\cite{shang2019end} and triplet classification~\cite{wang2017knowledge}.

There is a tendency that more and more scoring functions are proposed to model the interactions between entities and relations in recent years~\cite{hao2017end,bordes2013translating,sun2019rotate,trouillon2016complex,yang2014embedding}. As a result, the performance of KG embeddings has greatly been improved. However, compared with scoring functions, fewer previous works focus on another aspect of KG embeddings: negative sampling. Negative sampling is essential because training KG embeddings is a process in which true triplets are given higher scores while false triplets are given lower scores~\cite{ahrabian2020structure}. However, there are only ground truth triplets in KGs so negative triplets are used to generate corresponding false triplets.

Most current KG embedding models use uniform sampling to generate false triplets~\cite{bordes2013translating,trouillon2016complex,yang2014embedding}. Uniform sampling is to randomly select candidate entities with equal probability, so it is simple and efficient but only samples from a fixed distribution. Several pioneering works attempt to draw negative samples from a dynamical distribution. IGAN~\cite{wang2018incorporating} and KBGAN~\cite{cai2018kbgan} introduce generative adversarial network to select high-quality negative samples, while NSCaching~\cite{zhang2019nscaching} utilizes cached-based mechanism to pay more attention to high-quality negative samples. Differently, SANS~\cite{ahrabian2020structure} absorbs graph structure information into negative sampling process and achieves higher performances. Although these works optimize the negative sampling  mechanism from different aspects, a common shortcoming of these models still exists: these models can only select pre-existing entities to construct the negative samples, which is a limitation in exploring harder negative triplets. 

Recently, the mixing operation~\cite{zhang2017mixup} has shown its potential in many fields~\cite{lee2020mix,zhang2020seqmix,yoon2021ssmix}. Some researchers try to use the mixing operation to create better negative samples and the results show that the generated negative samples make the models more distinguishable~\cite{huang2021mixgcf,kalantidis2020hard}. Considering that most negative triplets are easily distinguishable and cannot provide valuable information, we introduce MixKG, a new mechanism for negative sampling in KGs. The first step in MixKG is to filter out hard negative triplets and there are two criteria: a score function-based selector and a correct entity similarity selector. The first criterion is to take the negative triplets with high scores in the score function as hard samples, while the second criterion considers the negative entities that have higher similarity with the correct entity as hard samples. After selecting hard negative samples, MixKG then uses the mixing operation over these hard negatives to generate harder negative samples. Despite the simplicity and efficiency of our model, the performance shows the superiority of our model. Our contributions are as follows:
\begin{itemize}
	\item To the best knowledge of us, we are the first to construct negative triplets using non-existent entities in KGs.
	\item We develop two criteria for selecting hard negative samples: score function-based selector and correct entity similarity selector.
	\item We generate harder negatives by mixing paired the selected hard negatives to make the model more distinguishable.
	\item We conduct extensive experiments on different datasets and KG embedding models to demonstrate the efficiency of the proposed model from different aspects.
\end{itemize}

\section{Related Works}
\subsection{Negative Sampling}
Negative sampling aims to generate negative samples, so the model can be trained by distinguishing observed positive data from negative samples. The advantage of negative sampling is abandoning to compute the normalization constant through sampling, which improves efficiency and effectiveness. From the perspective of sampling type, negative sampling can be divided into two categories: fixed negative sampling and dynamic negative sampling.

\subsubsection{Fixed negative sampling} As a classic sampling strategy, the idea of fixed negative sampling is simple and intuitive. There are  two representative methods for fixed negative sampling: uniform sampling ~\cite{bordes2013translating} and Bernoulli sampling~\cite{wang2014knowledge}. However, due to the restriction to fixed sampling limitation, the fixed sampling strategy fails to form harder negative samples.
and suffer from the vanishing gradient~\cite{cai2018kbgan}.
\subsubsection{Dynamic negative sampling}
To solve the problems in fixed negative sampling, several pioneering works have been proposed to extend negative sampling from fixed distribution to dynamic distribution. KBGAN~\cite{cai2018kbgan} and  IGAN~\cite{wang2018incorporating} attempt to absorb generative adversarial network to generate high-quality negative triplets. However, both KBGAN and IGAN win effectiveness at the expense of instability and degeneracy~\cite{zhang2019nscaching}. To reduce the complexness of the model while obtaining high-quality negative samples, NScaching~\cite{zhang2019nscaching} uses a cache to store negative triplets. In addition, SANS~\cite{ahrabian2020structure} puts the graph structure information in KGs in the negative sampling and then dynamically selects negative samples only from the \emph{l}-hop neighborhood of the head or tail entity. Although these methods try to generate high-quality negative samples from different aspects, they have a common shortcoming: the negative samples are generated only with pre-existing entities. 
\subsection{Mixing Method}
Mixing~\cite{zhang2017mixup} is a data augmentation method, which generates new data by convex combinations of pairs of samples. Essentially, mixup encourages the model to behave linearly in-between training samples. Mixing methods have shown their superiority in many applications~\cite{lee2020mix,zhang2020seqmix,yoon2021ssmix}. Overall, there are two popular types of domains for mixup: mixup for supervised learning and mixup for negative samples.
\subsubsection{Mixing for supervised learning}
Mixup~\cite{zhang2017mixup} generates fictitious training samples and their associated labels by linear interpolation, and the experimental results show mixup is universally applicable to image, speech and tabular datasets. SSMix~\cite{yoon2021ssmix} and MixText~\cite{chen2020mixtext} extend the mixup operation to the domain of natural language processing through input and hidden space mixing respectively.
\subsubsection{Mixing for hard negative samples}
The core idea of mixing is to construct virtual samples by linear interpolation so some researchers use mixing for harder negative mining. MixGCF~\cite{huang2021mixgcf} uses positive mixing and hop mixing to obtain high-quality negative samples, which achieves a higher performance in the recommended scenario. MoCHi~\cite{huang2021mixgcf} shows that harder negative samples can be obtained by mixing between hard negative samples and labels.
\section{Mix for Harder negative triplets}
In this section, we introduce the proposed method in detail. To note that,
the proposed method for generating harder negative samples via mixing operation is model-agnostic, so it is a general framework and can be easily plugged into the existing KGE models. 

The core of the proposed model can be divided into two steps: first, selecting high-quality negative triplets from the sampled negative triplets; second, mixing the paired high-quality negative triplets and then generating novel harder negative triplets.

\subsection{Select hard negative triplets}
In KGs, there are only true triplets as positive samples, negative triplets can be obtained by replacing the head or tail entities in true triplets with other entities sampled from the entire entity set. However, as is mentioned in previous works~\cite{zhang2019nscaching,ahrabian2020structure}, many of the sampled negative triplets cannot provide discriminative information to help the model learn effective embeddings of entities and relations. Only a few negative samples help the model converge in a right direction, and these samples are hard negative samples, which are the raw material for mixing in the proposed method. 

Hard negative samples are the core of the proposed model, but how to define hard negative samples remains a problem. Several previous works~\cite{zhang2019nscaching,kalantidis2020hard} use the score function $f$ and take the negative triplets with high scores as hard negative samples, and we call this method Score Function based Hard Negative Samples~(HNS-SF). However, this definition only considers the negative samples in the negative set and ignores the help of true triplets. Therefore, we introduce a different definition for hard negative samples, which considers candidate entities that are more similar to the correct head or tail entity as hard negative samples. We name this kind of definition of hard negative samples Correct Entity Similarity based Hard Negative Samples~(HNS-CES).

Given a KGE model scoring function $f$, a positive triplet~$(h, r, t)$, the set of entities $\mathcal{E}$. We first randomly sample entities from $\mathcal{E}$ then get the sampled negative candidate entities $(t_1^{'}, t_2^{'},... t_M^{'})$, and the negative triplets are \{$(h,r,t_1^{'}), (h,r,t_2^{'}),... (h,r,t_M^{'})$\} by replacing the tail entity in $(h, r, t)$ with each element in $(t_1^{'}, t_2^{'},... t_M^{'})$.  

To note that, we take corrupting the tail entity as an example, and generating negative samples via corrupting the head entity is similar. The specific process of selecting negative hard samples under the two definitions above are as follows:

\subsubsection{Hard Negative Samples-Score Function}
For each triplet in \{$(h,r,t_1^{'}), (h,r,t_2^{'}),... (h,r,t_M^{'})$\}, we calculate the scores $s(t_m^{'})=f(h, r, t_m^{'})$ using scoring function $f$. Then we sort the scores \{$s(t_1^{'}), s(t_2^{'}),... s(t_M^{'})$\} in a descending order, 
the negative triplets with K-largest scores are hard negative samples.
\subsubsection{Hard Negative Samples-Correct Entity Similarity}
We calculate the dot product between the tail entity  and each candidate entity in $(t_1^{'}, t_2^{'},... t_M^{'})$, then get the similarity  $s(t_m^{'})=t * t_m^{'}$. The candidate entities with top-K largest $s(t_m^{'})$ and the head entity, relation in $(h, r, t)$ form hard negative triplets. 

To note that, the size of the selected hard negative triplets K is a hyperparameter, which will be discussed in more detail in Section 4.

\subsection{Mixing among these hard triplets}
After obtaining the hard negative triplets for mixing, we  randomly select paired hard negative triplets then mix the tail entities to generate new harder negative samples.
To be precise, suppose there are two negative triplets \{$(h,r,t_i^{'}), (h,r,t_j^{'})$\}, then the tail entity of the newly generated harder negative triplet is
\begin{equation}
\hat{t}_{i,j} = \alpha * t_i^{'} + (1 - \alpha) * t_j^{'} 
\label{mix_equation}
\end{equation}
where $\alpha$ is randomly drawn from $(0,1)$, then the generated harder negative triplet is $(h, r, \hat{t}_{i,j})$.

\begin{algorithm}[tb]
	\caption{MixKG: Mixing negative samples for KG embeddings}
	\label{alg:algorithm}
	\textbf{Input}: training set $\mathcal{S}=\{(h, r, t)\}$, entity set $\mathcal{E}$, relation set $\mathcal{R}$, embedding dimension $d$, scoring function $f$, negative sample size $M$, generated negative sample size $N$, the size of selected harder negative number K, the size of epoches E.
	\begin{algorithmic}[1] 
		\STATE initialize embeddings for each $e \in \mathcal{E}$ and $r \in \mathcal{R}$.\\
		\WHILE{epoch $<$ E}
		\STATE sample a mini-batch  $\mathcal{S}_{\text {batch }} \in \mathcal{S}$;
		\WHILE{$(h, r, t) \in \mathcal{S}_{\text {batch }}$}
		\STATE uniformly sample $M$ entities from $\mathcal{E}$ to form negative triplets $\{(h, r, t_{m}^{'}), m=1,2...M\}$
		\STATE using scoring function or correct entity similarity to select negative triplets with K highest scores $\{(h, r, t_{k}^{'}), k=1,2...K\}$
		\WHILE{n $<$ N}
		\STATE uniformly pick up two triplets from negative triplets $\{(h, r, t_{k}^{'}), k=1,2...K\}$ and mix the tail entities using Equation~\ref{mix_equation} to get new negative triplet;
		\ENDWHILE \\
		calculate loss functions using Equation~\ref{loss_function_1} or \ref{loss_function_2}, then update the embeddings of entites and relations via gradient descent
		\ENDWHILE 
		\ENDWHILE
	\end{algorithmic}
\end{algorithm}

\subsection{Loss function}
The last step of KGE is to construct a loss function to distinguish the positive and negative samples. Based on this, the embeddings of entities and relations are learned by forcing the positive triplets to have higher scores while lower scores for negative samples.  The loss functions of the present KGE models can be divided into two categories: the first one is the translational distance model:
\begin{equation}
 \mathcal{L} =-\log \sigma(\gamma-f(h,r,t)) \\ -\sum_{i=1}^{n}  \log \sigma(f(h_{i}^{\prime},r,{t}_{i}^{\prime})-\gamma) 
 \label{loss_function_1}
\end{equation}
and the second one is the semantic matching model:
\begin{equation}
\mathcal{L} =-\log \sigma(f(h,r,t)) \\ -\sum_{i=1}^{n}  \log \sigma(-f(h_{i}^{\prime},r,{t}_{i}^{\prime})) 
\label{loss_function_2}
\end{equation}
where $(h_{i}^{\prime},r,{t}_{i}^{\prime})$ is the generated negative samples for $(h,r,t)$ by randomly replacing $h$ or $t$, $f$ is the scoring function to model the interactions between entities and relations. In the translational distance model, $f(h,r,t)$ is supposed to be smaller than $f(h_{i}^{\prime},r,{t}_{i}^{\prime})$, while $f(h,r,t)$ should be larger than $f(h_{i}^{\prime},r,{t}_{i}^{\prime})$ in semantic matching model, $\sigma$ is the sigmoid function, $\gamma$ is a fixed margin,  and $n$ is the size of negative samples.

\section{Experiments}
In this section, we conduct detailed experiments to demonstrate the effectiveness of the proposed model from various aspects. Our experiments seek to answer the following research questions~(RQs):
\begin{itemize}
	\item \textbf{RQ1}: Can the proposed method generate harder negative samples and have higher link prediction performances compared with previous negative sampling algorithms?
	\item \textbf{RQ2}: Are hard negative samples necessary for mixing? 
	\item \textbf{RQ3}: How the number of selected hard negative samples for mixing affects the performances? 
	\item \textbf{RQ4}: Can more generated negative samples lead to better performance?
\end{itemize}
\subsection{Datasets}
We conduct experiments on two public KG datasets: FB15k-237~\cite{toutanova2015observed} and WN18RR~\cite{dettmers2018convolutional}. FB15k-237 is a subset of FB15k, which comes from FreeBase~\cite{bollacker2008freebase} that contains lots of real-world triplets. Similarly, WN18RR is a subset of WN18 from WordNet KB, which is a large lexical English database. The details of the two datasets are given in Table~\ref{dataset_information}. 
\begin{table}[h]
	\caption{Statistics of datasets}\smallskip
	\centering
	\resizebox{0.65\columnwidth}{!}{
		\smallskip
		\begin{tabular}{c|c|c}
			\hline
			Dataset & WN18RR & FB15k-237\\
			\hline
			\#entity    & 14541  & 40943       \\
			\hline
			\#relation& 11  & 237   \\ 
			\hline
			\#train & 86835 &272115 \\
			\hline
			\#valid & 3034&17535\\
			\hline
			\#test &3134&20466\\
			\hline
		\end{tabular}
	}
	\label{dataset_information}
\end{table}
\subsection{Evaluation Protocols}
Following previous works~\cite{zhang2019nscaching,ahrabian2020structure}, we utilize two standard metrics to evaluate the performances for link prediction: mean reciprocal ranking~(MRR) and Hits@10. Assuming $\mathcal{N}$ is th size of test triplets, MRR is the average of the reciprocal ranks $\frac{1}{\mathcal{N}} \sum_{i=1}^{\mathcal{N}} \frac{1}{\operatorname{rank}_{i}}$, where $\operatorname{rank}_{i}, i \in\{1, \ldots,\mathcal{N}\}$ denotes the ranking results. In addition, Hits@10 is computed via $\frac{1}{\mathcal{N}} \sum_{i=1}^{\mathcal{N}} \mathbf{I}\left(\operatorname{rank}_{i}<10)\right.$, where $\mathbf{I}(\cdot)$ is the indicator function. To be consistency with the previous and avoid underestimation, the results in our experiments are in a filtered setting, where all corrupted triplets in the dataset are removed.
\subsection{Selected score functions} 
As is mentioned above, the KGE models can be divided into two groups: translational distance models and semantic matching models. We choose two representative models in each of the groups. For translational distance model, we utilize TransE~\cite{bordes2013translating} and RotatE~\cite{sun2019rotate}, then select ComplEx~\cite{trouillon2016complex} and DistMult~\cite{yang2014embedding} for semantic matching model. The definitions of the four models are listed in Table~\ref{definition_score_function}.
\begin{table}[h]
	\caption{The definitions of the four models, where $\odot$ denotes the Hadamard product, 
		$\langle\cdot\rangle$ denotes dot product,
		$\bar{\cdot}$ denotes conjugate for complex vectors}\smallskip
	\centering
	\resizebox{0.9\columnwidth}{!}{
		\smallskip
		\begin{tabular}{c|c|c}
			\hline
			model & scoring functions & definition\\
			\hline
			translational & TransE  & $\|\mathbf{h}+\mathbf{r}-\mathbf{t}\|$       \\
			distance & RotatE  &   $\|\mathbf{h} \circ \mathbf{r}-\mathbf{t}\|$     \\
			\hline
			semantic & DistMult  &$\langle\mathbf{r}, \mathbf{h}, \mathbf{t}\rangle$     \\
			matching & ComplEx  & $\operatorname{Re}(\langle\mathbf{r}, \mathbf{h}, \overline{\mathbf{t}}\rangle)$      \\
			\hline
		\end{tabular}
	}
	\label{definition_score_function}
\end{table}
\subsection{Hyperparameter settings}
The optimizer in our experiments is Adam~\cite{kingma2014adam} and we fine-tune the hyperparameters on the validation dataset. The embeddings of entities and relations are uniformly initialized. In addition, we use grid search to find the best hyperparameters, and the ranges of hyperparameters are followings: batch size $\in$ \{128, 256, 512\}, negative sample set size $M$ $\in$ \{256, 512, 1024, 2048\}, fixed margin $\gamma$ $\in$ \{3, 6, 9\}, selected top hard negative sample size K $\in$ \{5, 30 , 50, 100\}.

\begin{flushleft}
	\begin{table*}[h]
		\centering
		\caption{Results of different negative sampling algorithms for link prediction on four scoring functions and two public datasets. In the table, HNM-SF indicates the hard negative samples for mixing are selected via score function, while HNM-CES denotes the selection criterion is correct entity similarity.  Bold numbers represent the state-of-the-art performances, and the underlined numbers mean the best performances of the baselines.}\smallskip\smallskip
		\begin{tabular}{cccccc} 
			\hline
			\multirow{2}{*}{Score functions} 
			& Dataset	& \multicolumn{2}{c}{FB15k-237}  & \multicolumn{2}{c}{WN18RR}  \\
			\cline{2-6}
			& Metrics	& MRR & Hits@10(\%)  & MRR & Hits@10(\%)  \\
			\hline 
			\multirow{8}{*}{TransE}
			&KBGAN & 0.2926 & 46.59 & 0.1808 & 43.24 \\
			&NSCaching& \underline{0.2993} & 47.64 & 0.2002 & 47.83 \\
			&Uniform & 0.2927 & 48.03 & 0.2022 & 49.63 \\
			&Uniform SANS& 0.2962 & 48.35 & 0.2254 & 51.15 \\
			&Uniform RW-SANS& 0.2981 & \underline{48.50} & \underline{0.2317} & \underline{53.41} \\
			&\textbf{HNM-SF} & 0.2983 & \textbf{50.29} & 0.2332 & \textbf{53.61} \\
			&\textbf{HNM-CES}& \textbf{0.3032} & 49.55 & \textbf{0.2386} & 52.97 \\
			&\textbf{Improvement} & +0.0039 &  +1.94 & +0.0069 & +0.2\\
			\hline
			
			\multirow{6}{*}{RotatE}
			&Uniform & 0.2946 & 47.85 & 0.4711 & 56.51 \\
			&Uniform SANS& 0.2985 & 48.22 & 0.4769 & 55.76 \\
			&Uniform RW-SANS& \underline{0.3003} & \underline{48.47} & \underline{\textbf{0.4796}} & \underline{\textbf{57.12}} \\
			&\textbf{HNM-SF} & \textbf{0.3292} & \textbf{51.75} & 0.4731 & 56.36 \\
			&\textbf{HNM-CES}& 0.3002 & 48.11 & 0.4753 & 56.37 \\
			&\textbf{Improvement} & +0.0289 & +3.28 & -0.0043 &  -0.75\\
			\hline
			\hline
			
			\multirow{8}{*}{DistMult}
			&KBGAN & 0.2272 & 39.91 & 0.2039 & 29.52 \\
			&NSCaching& \underline{0.2834} & \underline{45.56} & \underline{0.4128} & 45.45 \\
			&Uniform & 0.2537 & 40.26 & 0.3938 & \underline{52.86} \\
			&Uniform SANS& 0.2595 & 41.00 & 0.4025 & 44.74 \\
			&Uniform RW-SANS& 0.2621 & 41.46 & 0.4071 & 49.09 \\
			&\textbf{HNM-SF} & \textbf{0.3085}& \textbf{48.68} & \textbf{0.4420} & \textbf{53.22} \\
			&\textbf{HNM-CES}& 0.2832 & 45.15 & 0.4299 & 50.91 \\
			&\textbf{Improvement} & +0.0251 & +3.12 & +0.0287 &  +0.36\\
			\hline
			
			\multirow{8}{*}{ComplEx}
			&KBGAN & 0.1910 & 32.07 & 0.3180 & 35.51 \\
			&NSCaching& \underline{0.3021} & \underline{48.05} & 0.4463 & 50.89 \\
			&Uniform & 0.2715 & 43.13 & \underline{0.4506} & \underline{56.05} \\
			&Uniform SANS& 0.2721 & 43.21 & 0.3832 & 41.16 \\
			&Uniform RW-SANS& 0.2819 & 44.62 & 0.4247 & 46.38 \\
			&\textbf{HNM-SF} & \textbf{0.3160} & \textbf{50.43} & \textbf{0.4729} & \textbf{56.29} \\
			&\textbf{HNM-CES}& 0.2712 & 43.03 & 0.4543 & 53.46 \\
			&\textbf{Improvement} & +0.0139 &+2.38 & +0.0223 &  +0.24\\
			\hline
		\end{tabular}
		\label{tab::main_results}
	\end{table*}
\end{flushleft}

\subsection{Baselines}
We utilize the following negative sampling algorithms as baselines:
\begin{itemize}
    \item KBGAN~\cite{cai2018kbgan} first uniformly samples some entities to construct candidate negative triplets, then leverages one generator to pick up one high-quality negative triplet for training.
    \item NSCaching~\cite{zhang2019nscaching} constructs a cache to store high-quality negative samples, then selects negative samples from the cache and dynamically introduces new high-quality negative samples into the cache.
    \item Uniform\cite{sun2019rotate} uniformly selects candidate entities from the entire entity set to form negative triplets. 
    \item SANS~\cite{ahrabian2020structure} incorporates graph structure information into negative sampling, and picks up negative samples from the \emph{l}-hop neighborhood. 
    \item RW-SANS~\cite{ahrabian2020structure} is similar to SANS, and uses random walks of length \emph{l} to approximate the \emph{l}-hop neighborhood.
\end{itemize}
The results of TransE, RotatE and DistMult in Table~\ref{tab::main_results} are taken from ~\cite{ahrabian2020structure}, while the results of ComplEx for KBGAN, NSCaching are taken from~\cite{zhang2019nscaching}, and the results of ComplEx for Uniform, Uniform SANS, Uniform RW-SANS are our reproductions using codes in \cite{ahrabian2020structure}.   
\subsection{Results analysis (RQ1)}
The results of different negative sampling algorithms on four score functions are shown in Table~\ref{tab::main_results}.
From Table~\ref{tab::main_results}, we can draw a conclusion that both \emph{HNM-SF} and \emph{HNM-CES} can improve the performances. We can see average improvement values of 
0.0180 and 0.0115 in MRR for FB15k-237 dataset and WN18RR dataset, respectively, while the improvement values are 1.07, 1.27, 1.74, 1.33 in Hits@10 for TransE, RotatE, DistMult and ComplEx respectively.

Overall, the two methods beat almost all previous negative sampling algorithms whether on the semantic translational distance models or on the semantic matching models, which shows the superiority of the proposed methods in general.
\subsection{The importance of hard negative samples for mixing(RQ2)}
\begin{figure}[H]  
	\centering  
	\includegraphics[width=1.0\columnwidth]{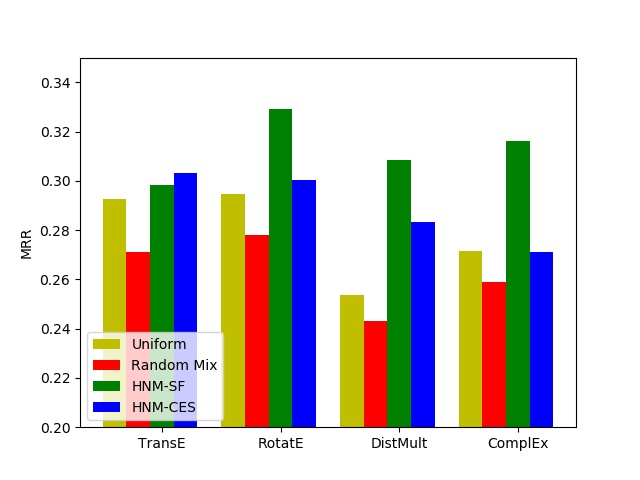}
	\caption{Comparsions of different mixing strategies on FB15k-237. Uniform denotes uniform sampling negative samples , Random Mix shows randomly picking up negative samples for mixing.} 
	\label{fig::zhuzhuangtu_237_MRR}  
\end{figure}

To validate whether selecting hard negative samples is necessary for mixing, we also conduct experiments under the setting of \emph{Random Mix}, which denotes the candidate entities for mixing are randomly drawn from the sampled negative samples without hard negative selection mechanism. The results are shown in Figure~\ref{fig::zhuzhuangtu_237_MRR}, and we have the following observations:
\begin{itemize}
	\item Compared with \emph{Uniform} on the four scoring functions, \emph{Random Mix} has lower performances, which  shows mixing among randomly drawn negative triplets cannot generate harder negatives.
	\item Both \emph{HNM-SF} and \emph{HNM-CES} improve \emph{Uniform} by a large margin, therefore we can conclude that only selecting high-quality negative samples can lead to harder negatives and performance improvements.
\end{itemize}
\subsection{How the size of hard negative samples affects performances (RQ3)}
\begin{figure}[H]  
	\centering  
	\includegraphics[width=1.0\columnwidth]{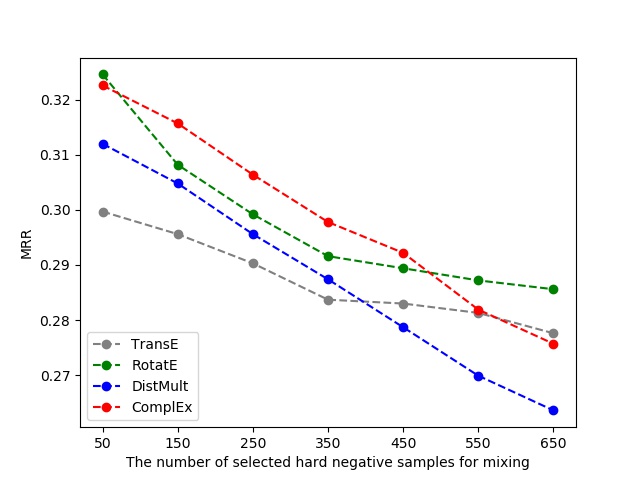}
	\caption{How the number of selected hard negative samples for mixing influence the performance on FB15k-237. The size of the negative sample set is 1024, and the x-axis denotes picking up how many hard negative samples from the 1024 sampled negative samples.} 
	\label{fig::zxt_237_negative_num_MRR}  
\end{figure}

We use score function or correct entity similarity to calculate the hard extent of the negative samples and select the top-K hard negatives as hard negative samples for mixing. 
To see how K affects the performances, we conduct various experiments with different K in a range of \{50, 150, 250, 350, 450, 550, 650\}. 
The results are shown in Figure~\ref{fig::zxt_237_negative_num_MRR}, we can have the observations:
\begin{itemize}
	\item The performances gradually decrease with the increasing number of the selected hard negative samples on all the scoring functions, which means that only a small number of negative samples are hard and helpful for guiding the KGE training process.
	\item With more unimportant negative samples joining the mixing process, it introduces noise and dilutes the proportion of the hard negative samples in final generated negative samples. Overall, the monotone decreasing line tells that the criteria for selecting hard negative samples are effective and essential.  
\end{itemize}
\subsection{How the size of generated harder negative samples affects performances (RQ4)}
In this section, we analyze how the number of generated harder negatives affect performances. The final harder negative samples are generated via mixing two hard negative samples that are randomly drawn from the selected hard negative set. Theoretically, an ocean of harder negative samples can be generated by drawing two hard negative samples and mixing them endlessly. 
Based on this, we  conduct detailed experiments on
how the number of generated negative samples $N$  affects the performances. 

\begin{figure}[H]  
	\centering  
	\includegraphics[width=1.0\columnwidth]{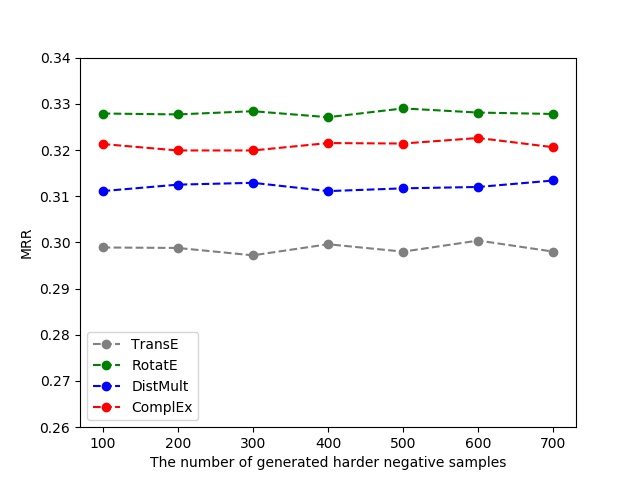}
	\caption{The relationship between the number of generated harder negative samples via mixing operation and MRR on the TransE, RotatE, DistMult and ComplEx for FB15k-237. } 
	\label{fig::zxt_237_generated_negative_num_MRR}  
\end{figure}

The results on the four score functions and FB15k-237 are in Figure~\ref{fig::zxt_237_generated_negative_num_MRR}, and the number of generated negatives is in range \{100, 200, 300, 400, 500, 600, 700\}. The results on four  different score functions all denote the performances remain consistency with a wide range number of generated hard negative samples, which shows that more generated negatives cannot improve performances. In other words, it is the generated high-quality harder negative samples which can benefit the performance not more generated negative samples, since more generated hard negative samples cannot provide extra effective information for training.

\section{Conclusion}
In this paper, we consider constructing harder negative samples via mixing hard negative samples, which is a simple but powerful method and can be absorbed into the present KGE models easily.  Our work sheds light on the effectiveness and importance of generating non-existent entities as negative samples. We develop two kinds of criteria for picking up hard negative samples.
one is taking the negative triplets with high scores as negative triplets, the other is selecting the candidate negative entities similar to a correct entity as negative samples, and the two types of hard negative samples can both lead to harder negative samples via mixing operation.
We test the proposed method on four score functions and two public datasets, and the results show the method owns generalization under various experimental settings. There can be other sides of information to select hard negative samples for mixing, such as graph structure information~\cite{ahrabian2020structure}, and we leave this for future work.

\bibliographystyle{named}
\bibliography{ijcai22}

\end{document}